\documentclass[runningheads]{llncs}
\usepackage{graphicx}
\usepackage{amsmath,amssymb} 
\usepackage{arydshln}
\usepackage{array}

\usepackage{times}
\usepackage{epsfig}
\usepackage{multirow} 
\usepackage{booktabs} 
\usepackage{algorithm}
\usepackage[noend]{algpseudocode}
\usepackage{makecell}
\usepackage{lipsum}
\usepackage{listings}
\usepackage{caption}
\usepackage{wrapfig}
\usepackage[table,x11names]{xcolor}

\newcommand{\trimmedgraphic}[2][]{%
	\includegraphics[trim = 6.5cm 9cm 5cm 9cm,clip,scale=0.18,#1]%
	{#2}%
}

\newcommand{\trimmedfiedler}[2][]{%
	\includegraphics[trim = 5cm 8cm 5cm 8cm,clip,scale=0.17,#1]%
	{#2}%
}

\newcommand{\etal}{\textit{et al}. }

\usepackage{hyperref}

\usepackage[width=122mm,left=12mm,paperwidth=146mm,height=193mm,top=12mm,paperheight=217mm]{geometry}

\begin{document}

\pagestyle{headings}
\mainmatter

\title{Local Spectral Graph Convolution for Point Set Feature Learning} 


\authorrunning{C. Wang \etal}

\author{Chu Wang, Babak Samari, Kaleem Siddiqi}


\institute{School of Computer Science and Center for Intelligent Machines,\\
	McGill University\\
	\email{ \{chuwang,babak,siddiqi\}@cim.mcgill.ca}
}

\maketitle

\begin{abstract}
Feature learning on point clouds has shown great promise, with the introduction of effective and generalizable deep learning frameworks such as pointnet++. Thus far, however, point features have been abstracted in an independent and isolated manner, ignoring the relative layout of neighboring points as well as their features. In the present article, we propose to overcome this limitation by using spectral graph convolution on a local graph, combined with a novel graph pooling strategy. In our approach, graph convolution is carried out on a nearest neighbor graph constructed from a point's neighborhood, such that features are jointly learned. We replace the standard max pooling step with a recursive clustering and pooling strategy, devised to aggregate information from within clusters of nodes that are close to one another in their spectral coordinates, leading to richer overall feature descriptors. Through extensive experiments on diverse datasets, we show a consistent demonstrable advantage for the tasks of both point set classification and segmentation.

\end{abstract}

\section{Introduction}

With the present availability of registered depth and appearance images of complex real-world scenes, there is tremendous interest in feature processing algorithms for classic computer vision problems including object detection, classification and segmentation. In their latest incarnation, for example, depth sensors are now found in the apple iPhone X camera, making a whole new range of computer vision technology available to the common user. For such data it is particularly attractive to work {\em directly} with the unorganized 3D point clouds and to not require an intermediate representation such as a surface mesh. The processing of 3D point clouds from such sensors remains challenging, since the sensed depth points can vary in spatial density, can be incomplete due to occlusion or perspective effects and can suffer from sensor noise. 

Motivated by the need to handle unstructured 3D point clouds while leveraging the power of deep neural networks, the pointnet++ framework has shown promise for 3D point cloud feature processing for the tasks of recognition and segmentation \cite{qi2017pointnet}. In this approach a network structure is designed to work directly with point cloud data, while aggregating information in a hierarchical fashion, in the spirit of traditional CNNs on regular grids. To do so, a centroid sampling is first applied on the input point cloud, followed by a radius search to form point neighborhoods. Then the point neighborhoods are processed by multi-layer perceptrons \cite{qi2016pointnet} and the resulting point features are abstracted by a pooling operation. Through hierarchical multi-layer learning on the point cloud data, the pointnet++ framework exhibits impressive performance in both segmentation and classification on challenging benchmarks, while treating the input data as an unorganized point cloud.

In a parallel development, Defferrard \etal have sought to extend CNNs, traditionally applied on regular domains, such as
sampled image pixels in 2D or voxels in 3D, to irregular domains represented as graphs \cite{defferrard2016convolutional}. Their approach uses Chebyshev polynomials to approximate spectral graph filters; an initial graph is processed by convolutional operations to yield features which are then coarsened using sub-sampling and pooling methods. Kipf and Welling \cite{kipf2016semi} simplify the higher order polynomial approximations in Defferrard \etal and propose a first order linear approximation of spectral graph filters. The aforementioned spectral approaches operate on the full graph and have the limitation that the graph Laplacian and the graph coarsening hierarchy have to be precomputed, in an offline manner, before the network training or testing. This adds significant overhead when the full graph is large.

\begin{figure}[!t]
  \vspace{0.3cm}
  \centerline{
  \includegraphics[scale=0.57]{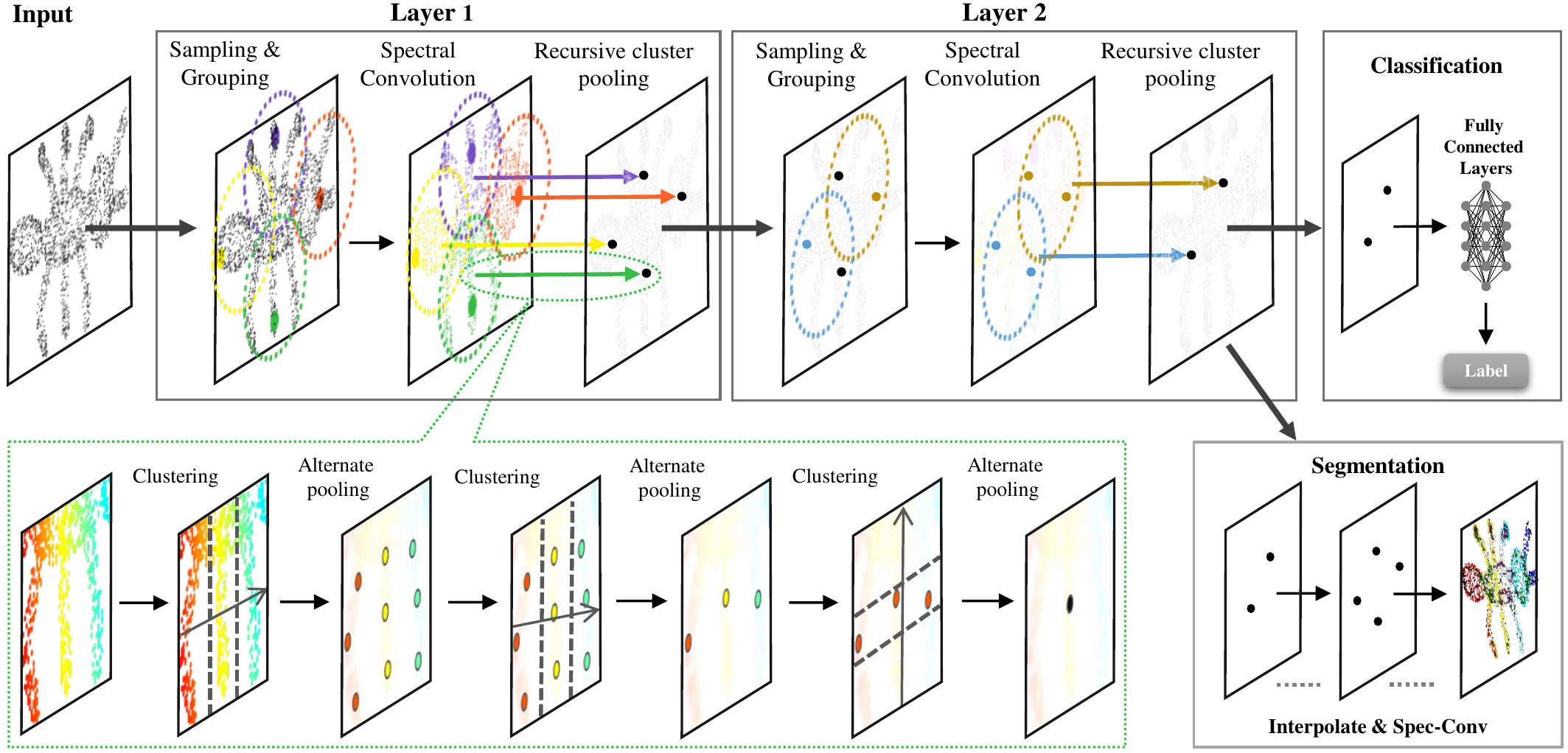}
  }
  \vspace{0.1cm}
  \caption{{\sc Top:} Starting from a point cloud, farthest point sampling leads to centroids, from which $k$-NN's are sampled. Then, for each neighborhood, spectral convolution is carried out followed by recursive cluster pooling. 
After several layers of sampling, spectral convolution and cluster pooling, we perform segmentation or classification.
{\sc Bottom:} The green dashed box details the process of recursive spectral cluster pooling on the Fiedler vector of a sample neighborhood. See text in Section \ref{sec:specpooling} for a discussion.
}
\label{fig:spectralpooling}
\end{figure}

In this article we propose to leverage the power of spectral graph CNNs in the pointnet++ framework, while adopting a different pooling strategy. This allows us to address two limitations of present deep learning methods from point clouds: 1) the fact that for each point sample the learning of features is carried out in an isolated manner in a local neighborhood and 2) that the aggregation of information in later layers uses a greedy winner-take-all max pooling strategy. Instead, we adopt a different pooling module, as illustrated by the detailed example in Fig. \ref{fig:spectralpooling}. 
Further, our method requires no precomputation, in contrast to existing spectral graph CNN approaches \cite{defferrard2016convolutional,kipf2016semi}. Our combination of local spectral feature learning with recursive clustering and pooling provides a novel architecture for point set feature abstraction from unorganized point clouds. Our main methodological contributions are the following:
\begin{itemize}
\item The use of local spectral graph convolution in point set feature learning to incorporate structural information in the neighborhood of each point.
\item An implementation of the local spectral graph convolution layer that requires no offline computation and is trainable in an end-to-end manner. We build the graph dynamically during runtime and compute the Laplacian and pooling hierarchy on the fly. 
\item The use of a novel and effective graph pooling strategy, which aggregates features at graph nodes by recursively clustering the spectral coordinates. 
\end{itemize}
The proposed architecture leads to new state-of-the-art object recognition and segmentation results on diverse datasets, as demonstrated by extensive experiments.

\section{Challenges in point set feature learning}\label{sec:limPointnet}
A limitation of feature learning in the pointnet++ framework \cite{qi2017pointnet}, is that features from the $k$ nearest neighbors ($k$-NN) of a point are learned in an isolated fashion. Let $h$ represent the output of an arbitrary hidden layer in a deep network, typically a multilayer perceptron. In pointnet++ the individual features for each point in the $k$-NN are achieved with $h(x_i), i \in {1,2,...,k}.$ Unfortunately, this hidden layer function does not model the joint relationship between points in the $k$-NN. A convolution kernel that jointly learns features from all points in the $k$-NN would capture topological information related to the geometric layout of the points, as well as features related to the input point samples themselves, e.g., color, texture, or other attributes. In the following section we shall extend approaches such as the pointnet++ framework to achieve this goal by using local graph convolution, but in the spectral domain. 

Another limitation in pointnet++ is that the set activation function for the $k$-NN is achieved by max pooling across the hidden layer's output for each point, such that
\begin{equation}
f(x_1,x_2, ... , x_k) = \max_{i \in {1,...,k}}  h(x_i).
\end{equation}
Max pooling does not allow for the preservation of information from disjoint sets of points within the neighborhood, as in the case of the legs of the ant in Fig. \ref{fig:spectralpooling}.
To address this limitation we introduce a recursive spectral clustering and pooling module that yields an improved set activation function for the $k$-NN, as discussed in Section (\ref{sec:specpooling}). The combined point set feature abstraction operation in this paper can be summarized by
\begin{equation}
f(x_1,x_2, ... , x_k) = \mathcal{\Phi}(h_1,h_2,...,h_k),
\end{equation}
where $h_i$ is the convolution output $h(x_1,x_2,...,x_k)$ evaluated at the $i$-th point and $\mathcal{\Phi}$ stands for our proposed set activation function. 

\begin{figure}[t]
	\centering
	\includegraphics[scale=0.32]{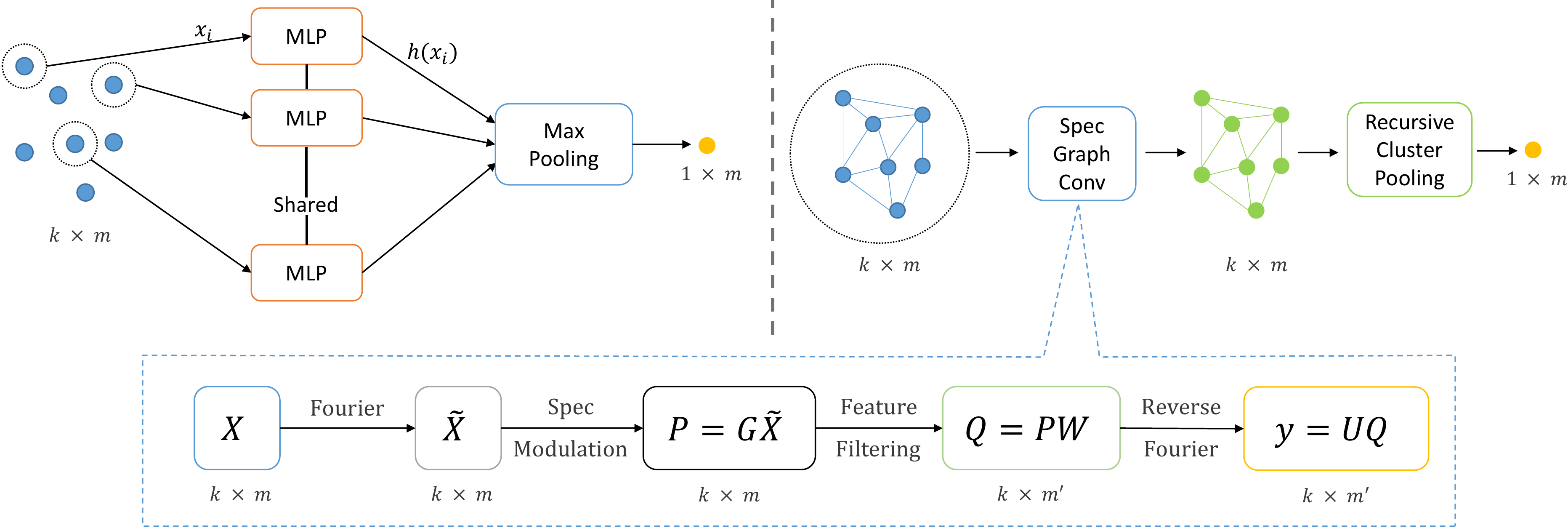}
	\caption{{\sc Top:} A comparison between point-wise MLP in pointnet++ (left) and our spectral graph convolution (right) in a local neighborhood. For each point, the spectral graph convolution output depends on all points in its neighborhood, whereas in point-wise MLP the output only depends on the point itself. {\sc Bottom:} We illustrate the network operations in a spectral graph convolutional layer, with the corresponding input/output dimensions.}
	\label{fig:fig0}
\end{figure}

Fig. \ref{fig:fig0} provides a comparison between the point-wise MLP in pointnet++ \cite{qi2017pointnet} and our spectral graph convolution, to better illustrate our motivation. Whereas pointnet++ abstracts point features in an isolated manner, spectral graph convolution considers all points in a local neighborhood in a joint manner, incorporating both features at neighboring points as well as structural information encoded in the graph topology in the abstraction. More formally, this is accomplished via the graph Fourier transform and spectral modulation steps, which blend neighborhood features using the eigenspace of the graph Laplacian (see Fig. \ref{fig:fiedler}). 
In the following section, we provide theoretical background and implementation details of our spectral graph convolution kernel.

\section{Graph Convolution}\label{sec:gCNN}
The convolution operation in the spatial domain (directly on vertices in the graph) is described by
\begin{equation}
h = X * g,
\end{equation}
where $X$ stands for the input point set features and $g$ for a spatial convolution kernel. This is equivalent to an element-wise Hadamard product in the graph spectral domain, as is shown in Defferrard \etal \cite{defferrard2016convolutional} and Shuman \etal \cite{shuman2013emerging}
\begin{equation}
\tilde h = \tilde X \odot \tilde g .
\end{equation}
Here $\tilde{X}$ stands for the graph Fourier transform of the point set features, $\tilde{g}$ stands for the filter in the graph Fourier domain and $\tilde{h}$ for the filtered output. In order to acquire the filtered output in the original spatial (vertex) domain, an inverse Fourier transform is required. We elaborate on the graph Fourier transform and the spectrum filtering below.

\subsection{Graph Formulation of a Local Neighborhood}
Given a set of $k$ points ${x_1, x_2, ..., x_k}$ in a local neighborhood, we build a representation graph $G_{k}$ whose vertices $V$ are the points and whose edges $E \subseteq V \times V$ carry weights $w : E \rightarrow \mathbb{R}_+^*$ based on a measurement of pair-wise distance, such as Euclidean distance between $xyz$ spatial coordinates or distance in a feature space provided by the deep network. This provides a graph adjacency matrix $W$, which is $k \times k$ nonnegative, symmetric, with entries $W_{ij} = dist(x_i,x_j)$. We then compute the graph spectrum based on this adjacency matrix and perform a graph Fourier transform, spectral filtering and finally an inverse Fourier transform.

\subsection{Graph Fourier Transform}
To compute a graph Fourier Transform of the point features $X \in \mathbb{R}^{k \times m}$, which are graph signals on vertices of $G_{k}$, we first need to compute the normalized graph Laplacian defined as
\begin{equation}\label{eqn:laplacian}
{\cal L} = I - D^{−1/2} W D^{−1/2},
\end{equation}
where $I$ is the identity matrix and $D \in  \mathbb{R}^{k \times k}$ is the diagonal degree matrix with entries $D_{ii} = \sum_{j} W_{ij} $.
It follows that ${\cal L}$ is a real symmetric positive semidefinite matrix, and has a complete set of orthonormal eigenvectors which comprise the graph Fourier basis 
\begin{equation}
U = [u_0,u_1,...,u_{k-1}] \in \mathbb{R}^{k \times k}.
\end{equation}
The eigenvalues can be used to construct a diagonal matrix 
\begin{equation}
\Lambda = diag([\lambda_0, \lambda_1, ..., \lambda_{k-1}]) \in \mathbb{R}^{k \times k}
\end{equation}
which contains the frequencies of the graph. Then it follows that ${\cal L} = U \Lambda U^{T}$. 
The graph Fourier transform of $X$ is then defined as
$\tilde{X} = U^T X$
and its inverse as 
$X = U \tilde{X}$.

\subsection{Spectral Filtering}

The convolution operation is defined in the Fourier domain as
\begin{equation}
x * g = U ((U^T x) \odot (U^T g)),
\end{equation} 
following Shuman \etal \cite{shuman2013emerging}, where $\odot $ is the element-wise Hadamard product, $x$ is an arbitrary graph signal and $g$ is a spatial filter. If we define $y = x * g$ as the output of the graph convolution, it follows that a graph signal $X \in \mathbb{R}^{k \times m} $ filtered by $g$ can be written as
\begin{align}\label{eqn:gconv}
y & = \tilde{g}_{\theta}(\mathcal{L}) X = \tilde{g}_{\theta}( U \Lambda U^T)  X =  U \tilde{g}_{\theta}(\Lambda) \tilde{X},
\end{align}
where $\theta$ stands for an arbitrary parametrization. In the following section, we describe our implementation of spectral filtering, which is introduced as a module on top of the existing pointnet++ \cite{qi2017pointnet} framework, using TensorFlow.

\subsection{Implementation of Spectral Filtering}
We carry out spectral graph convolution using standard unparametrized Fourier kernels, where the entries of $\tilde{g}_{\theta}(\Lambda)$ are all learnable. With $m$ the input feature dimension and $m'$ the output dimension, convolution of a graph signal $X \in \mathbb{R}^{k \times m}$ with spectral filters can be achieved by the following three steps:
\begin{enumerate}
\item \textbf{Spectral modulation} which outputs $P = G \tilde{X}$, with the diagonal matrix $G$ being the unparametrized kernel $\tilde{g}_{\theta}(\Lambda)$. The $k$ diagonal entries of $G$ are all free parameters in the unparametrized Fourier kernel formulation.
\item \textbf{Feature filtering} which expands the input dimension from $m$ to $m'$. The output of this step is a feature matrix $Q \in  \mathbb{R}^{k \times m'} $. The entry $q_{k,i}$ is the $i$-th output feature of the $k$-th point and is given by $q_{k,i} = \sum_{j = 1}^m p_{k , j} w_{j , i}.$
Here $p_{k , j}$ is the entry of $P$ corresponding to the $j$-th input feature of the $k$-th point defined in the previous step and $w_{j,i}$ is the filter coefficient between the $i$-th input feature with $j$-th output feature. This step can be represented by $Q = P W$,
where $W$ is the matrix of learnable filter parameters.
The filtering operation in steps 1 and 2 can be summarized as 
\begin{equation}
Q = (G \tilde{X}) W.
\end{equation}

\item \textbf{Reverse Fourier transform} which provides convolution outputs in the spatial graph signal domain via $y = U Q $.
\end{enumerate}
The above formulation resembles that of \cite{defferrard2016convolutional} and \cite{kipf2016semi}, with the difference that we build the $k$-NN graph during runtime, computing its Laplacian and pooling hierarchy on the fly, thereby requiring no offline precomputation. We further note that the weights of the feature filter $W$, as well as the spectral modulation matrix $G$, are shared by all the different local neighborhoods in a given graph convolution layer. Thus, unlike \cite{defferrard2016convolutional,kipf2016semi}, the learned parameters in our work do not depend on the underlying graph structure. Fig. \ref{fig:fig0} (bottom) illustrates the above spectral filtering process.

While the more sophisticated efficient kernels of \cite{defferrard2016convolutional} could be used, our goal was to demonstrate the improvement obtained by graph CNNs in general. The overhead of eigenvalue decomposition (EVD) in our unparametrized spectral kernel does not significantly affect runtime since the EVD is computed on local $k$-NN graphs, with $k$ being very small. This computation is easily handled by parallel computing on GPUs as demonstrated by the experiments showing training time cost in our model ablation study in Section \ref{sec:modelAblation}.

\section{Pooling on Local $k$-NN Graph}\label{sec:specpooling}
The set activation function discussed in Section \ref{sec:limPointnet}, whose aim is to summarize information from a $k$-NN graph, is essentially a form of graph pooling, where a graph of $k$ vertices is abstracted via feature pooling to a single vertex. We propose a novel $k$-NN graph pooling algorithm using hierarchical clustering and within-cluster pooling.

The general strategy is to pool information during learning in a manner that does so only {\em within} a cluster of similar abstract point features, in contrast to the greedy strategy of max pooling, which is commonly applied in regular CNNs as well as in pointnet++. The intuition here is that multiple sets of distinct features may together contribute towards a salient property for the task at hand, and that the detection of these clusters combined with within cluster pooling will improve performance. For example, for the task of classifying a point set sampled from a human head one would want to simultaneously learn and capture nose-like, chin-like and ear-like features and not assume that only one of these would be discriminative for this object category. We discuss each of our steps in turn. 

\subsection{Spectral Clustering}\label{sec:eigL}
We propose to group features into clusters according to a local $k$-NN's geometric information that is embedded using spectral coordinates. The spectral coordinates we use are based on the low frequency eigenvectors of the Graph Laplacian $\mathcal{L}$, which capture coarse shape properties since the Laplacian itself encodes pairwise distances in the $k$-NN graph. As exploited in \cite{lombaert2013focusr} \cite{bronstein2008book} for computing point-to-point correspondences on meshes or for feature description, the low frequency spectral coordinates provide a discriminative spectral embedding of an object's local geometry (see Fig. \ref{fig:spectralpooling} bottom and Fig. \ref{fig:fiedler}). The eigenvector corresponding to the second smallest eigenvalue, the Fiedler vector \cite{chung1997spectral}, is widely used for spectral clustering \cite{shi2000normalized}.

\begin{table}[t]
	\begin{center}
		\begin{tabular}{c c c c c c}
			& \textbf{bird}& & &\textbf{human}&\\
			\includegraphics[trim = 5cm 8cm 5cm 8cm,clip,scale=0.18]{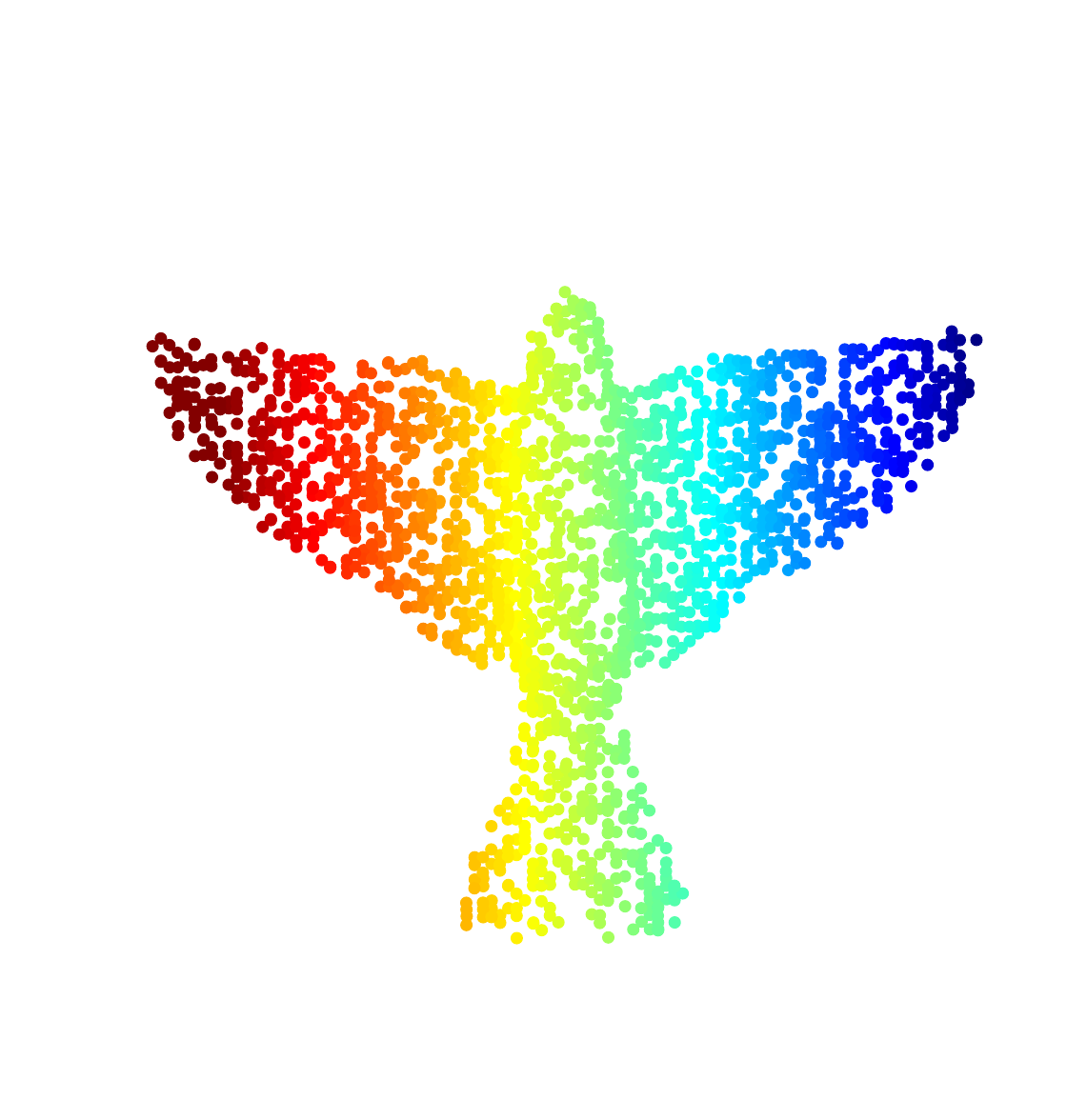} &			\trimmedfiedler{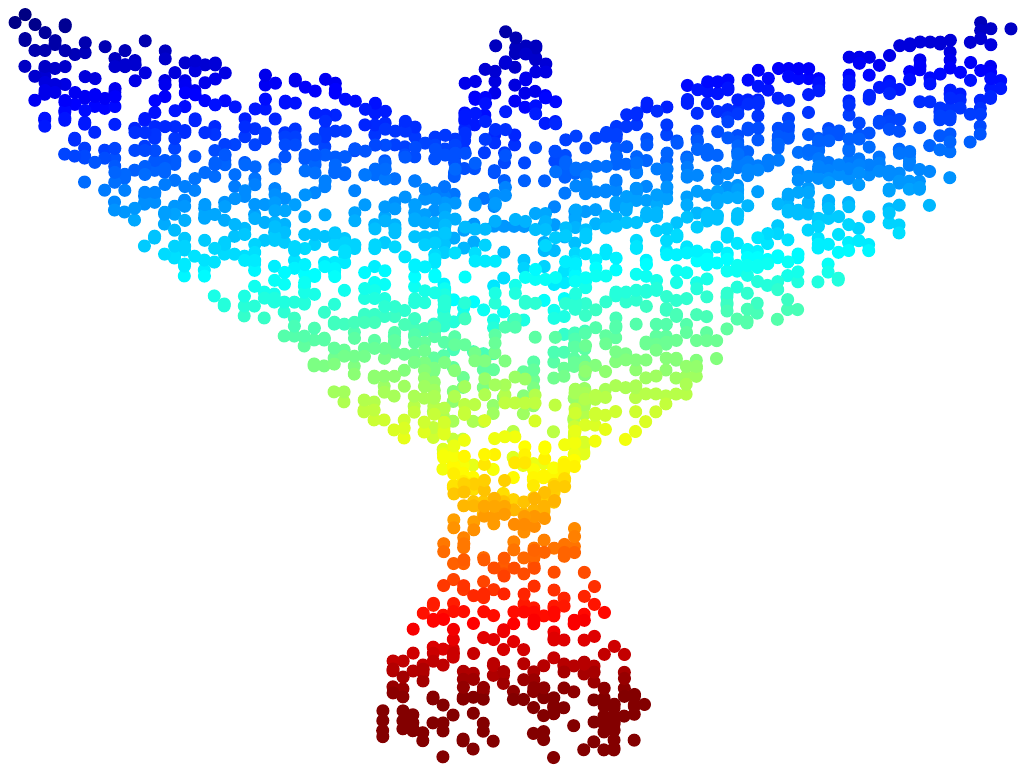} &			\includegraphics[trim = 5cm 8cm 5cm 8cm,clip,scale=0.18]{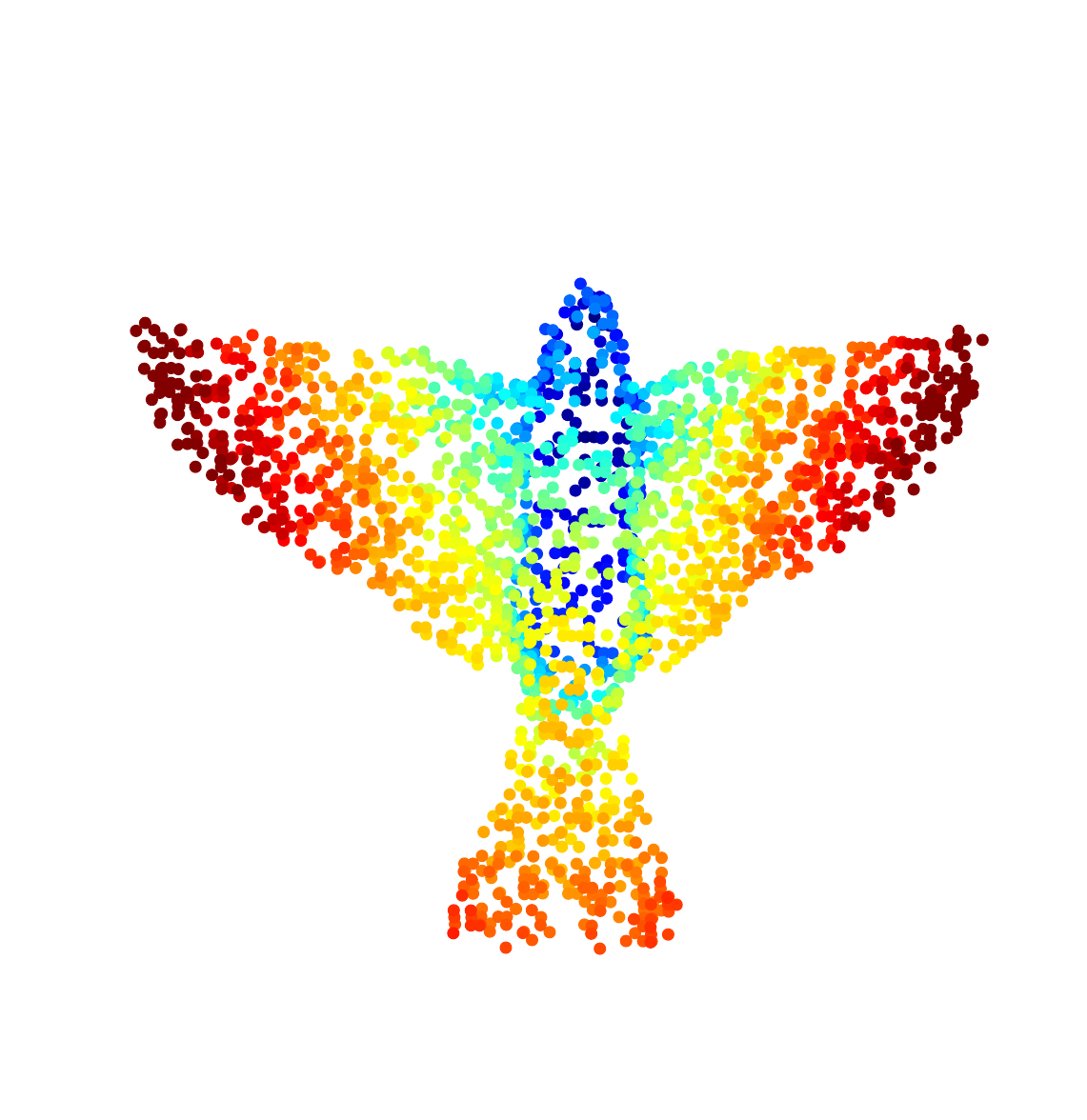} &	
			
			\trimmedfiedler{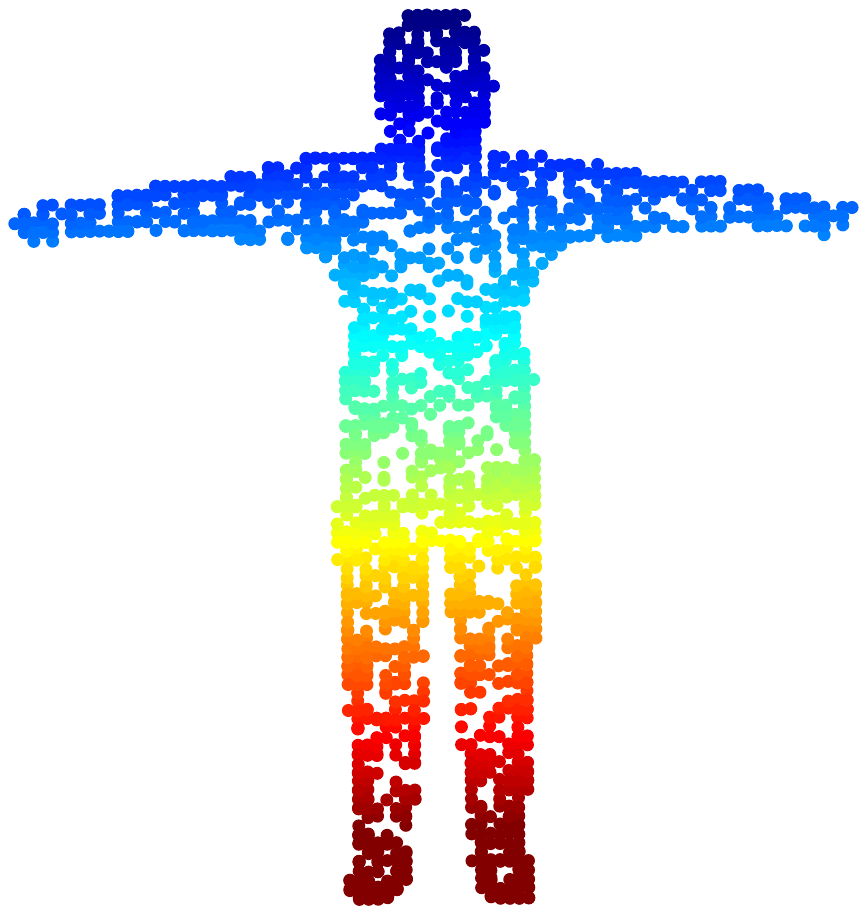} &		\trimmedfiedler{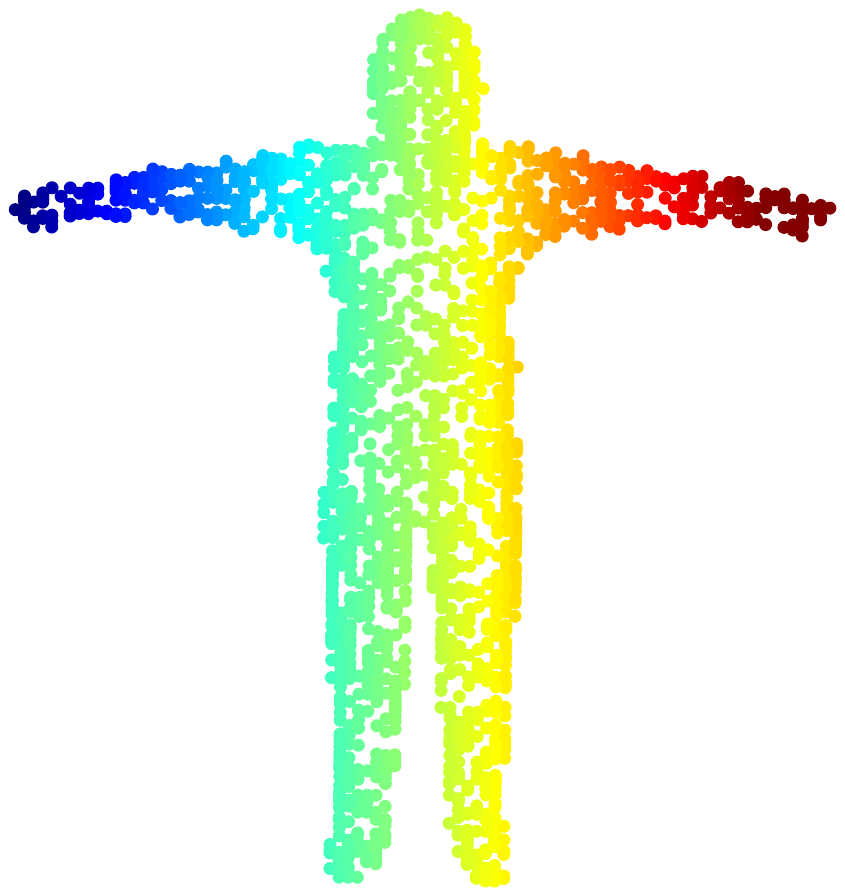} &			\trimmedfiedler{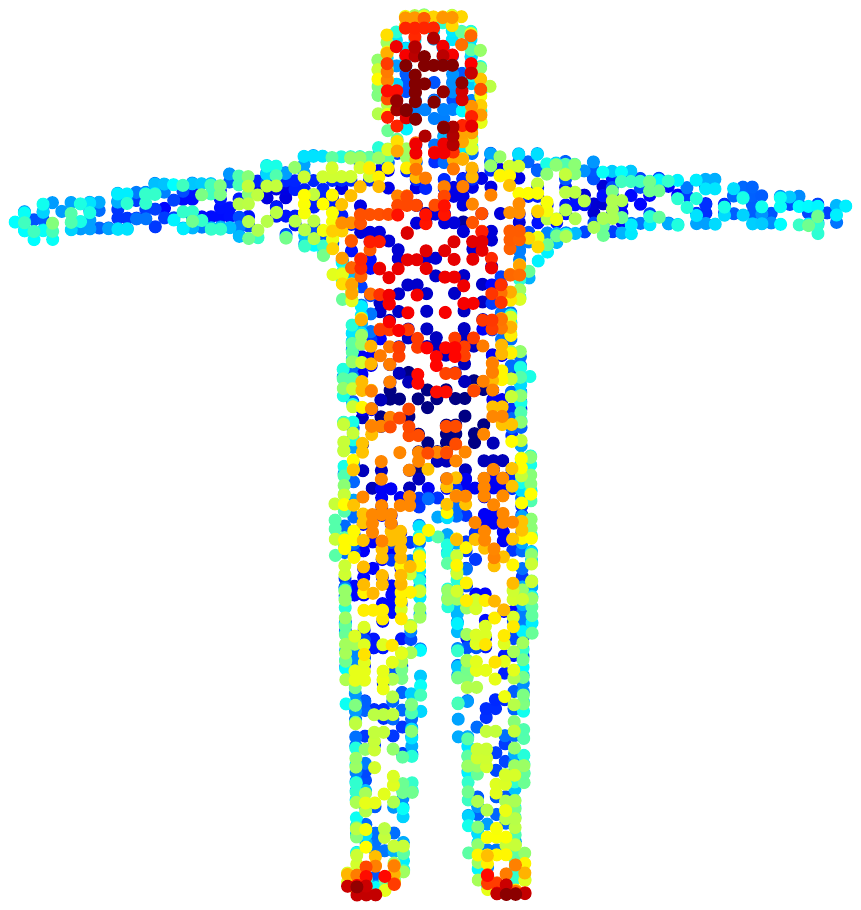}\\ 			
			
			$\lambda_1$&$\lambda_2$&$\lambda_3$&$\lambda_1$&$\lambda_2 $&$\lambda_3$\\
		\end{tabular}
	\end{center}
	\vspace{-0.1cm}
	\captionof{figure}{A visualization of spectral coordinates for models of a bird and a human, both from the McGill Shape Benchmark. $\lambda_i,i\in \{0,1, ..., k-1\}$ is the $i$-th eigenvalue of the graph Laplacian.}
	\label{fig:fiedler}
	\vspace{-0.8cm}
\end{table}

\subsection{Clustering, Pooling and Recurrences}\label{sec:clustering}
Following the normalized cuts clustering algorithm \cite{shi2000normalized}, we partition the Fiedler vector to perform spectral clustering in a local $k$-NN. We first sort the entries of the Fiedler vector in ascending order of their numerical value, and then evenly cut it into $k_1$ sections in the first iteration. The points in the neighborhood whose Fiedler vector entries fall in the same section will be clustered together. This results in $k_1$ clusters with a cluster size of $c = \frac{k}{k_1}$. After obtaining a partition into $k_1$ clusters, we perform pooling operations only {\em within} each cluster. This allows the network to take advantage of separated features from possibly disjoint components in the $k$-NN graph. 

The above steps result in a coarsened $k_1$-NN graph with aggregated point features. The same process is then repeated on the coarsened graph in a recursive manner, to obtain $k_i$ clusters for each iteration $i$. Note that we alternate between max pooling and average pooling between different recurrences to further increase the discriminative power of the graph pooling algorithm. The proposed algorithm terminates when the number of vertices remaining is smaller or equal to a prescribed cluster size. A regular full stride pooling is then applied on the resultant graph signals. We formalize the above steps in Algorithm \ref{algorithm1}. 

In practice, we found that using $k = 2 c^2$ as a relationship between cluster size and neighborhood size gave good results, with two recurrences of cluster pooling and a final pooling of size 2. We used max pooling as the first stage in the alternating pooling scheme, and fixed these configurations for all our experiments in Section \ref{sec:exp}. We implemented the recursive cluster pooling module in TensorFlow, integrating it fully with the spectral graph convolution layer to make the resultant network end-to-end trainable.

\medskip
\begin{algorithm}[t]
	\small
	\caption{Recursive Cluster Pooling}
	\label{algorithm1}
	\hspace*{\algorithmicindent}\textbf{\textsc{Inputs}}: pts $\gets$ point features $(\mathbb{R}^{k \times m})$, csize $\gets$ cluster size, and \textsc{Pool} $\gets$ the Pool method.\\
	\hspace*{\algorithmicindent}\textbf{\textsc{Output}}: Pooled point features $(\mathbb{R}^{1 \times m})$.\\
	\hspace*{\algorithmicindent}\textbf{\textsc{Methods}}: \textsc{Arg\_Sort}(x) returns the sorted indices of x, \textsc{ReArrenge}(x,y) permutes x along its 1st dimension according to the given indices y,  \textsc{Pool}(x) pools x along its 1st dimension.

	\begin{multicols}{2}
		
		\begin{algorithmic}
			\Procedure{Cluster}{pts, csize, \textsc{Pool}}
			\State $G\gets$ \textsc{Adj\_Matrix}(pts)
			\State $L\gets$ \textsc{Laplacian($G$)}
			\State $[\Lambda, U]\gets$ \textsc{EVD}($L$)
			\State fiedler\_vector$\gets U[:,1]$ 
			\State inds$\gets$ \textsc{Arg\_Sort}(fiedler\_vector)
			\State \textsc{ReArrenge}(pts, inds)
			\State \textsc{ReShape}(pts, [:, csize])
			\State \textbf{return} \textsc{Pool}(pts)
			\EndProcedure

			\Procedure{Main}{pts, csize}
			\State \textsc{Pool}$\gets$ \textsc{Max\_Pool}
			\While{\textsc{Count}(pts)$>$csize}
			\State pts$\gets$\textsc{Cluster}(pts, csize, \textsc{Pool})
			\If{\textsc{Pool} $==$ \textsc{Max\_Pool}}
			\State \textsc{Pool}$\gets$ \textsc{Avg\_Pool}
			\Else
			\State \textsc{Pool}$\gets$ \textsc{Max\_Pool}
			\EndIf
			\EndWhile
			\State \textbf{return}  \textsc{Pool}(pts)
			\EndProcedure
			
		\end{algorithmic}
	\end{multicols}
\end{algorithm}

\section{Experiments}\label{sec:exp}
\subsection{Datasets}\label{sec:datasets}

We evaluate our approach against present state-of-the-art methods on the following 5 datasets:
\begin{itemize}
\item MNIST: This contains images of handwritten digits with 60k training and 10k testing samples. It has been used to benchmark related graph CNN approaches \cite{defferrard2016convolutional,monti2016geometric} as well as pointnet++ \cite{qi2017pointnet}.

\item ModelNet40 \cite{wu20153d}: This contains CAD models of 40 categories, sampled into point clouds. We use the official split, with 9,843 training and 2,468 testing examples.

\item McGill Shape Benchmark \cite{siddiqi2008retrieval}: This dataset contains 456 CAD models of 19 object level categories. We sample the meshes into point clouds and use the first two-thirds of the examples in each category for training and the remaining one-third for testing. The dataset is also divided into articulated (254 models) versus non-articulated (202 models) sub-categories.

\item ShapeNet part segmentation dataset \cite{yi2016scalable}: This dataset contains 16,881 shapes from 16 classes, with the points of each model labeled into one of 50 part types. We use the official training/testing split, following \cite{qi2016pointnet,yi2016scalable,yi2016syncspeccnn}, where the challenge is to assign a part label to each point in the test set.

\item ScanNet Indoor Scene dataset \cite{dai2017scannet}: This dataset contains 1513 scanned and reconstructed indoor scenes, with rich annotations including semantic voxel labels. We follow the experimental settings for segmentation in \cite{qi2017pointnet,dai2017scannet} and use 1201 scenes for training, and 312 scenes for testing.
\end{itemize}

\begin{table}[!ht]
	\vspace{-0.4cm}
	\parbox{.5\linewidth}{
		\centering
		\small
		\renewcommand{\arraystretch}{1.2}
		\scalebox{0.80}{
			\begin{tabular}{l c | c c c c |c|c}
				\hline
				
				\multicolumn{2}{l|}{Structure}	& L1  			& L2		   & L3 		& L4	&	Kernel    & Pooling 		  \\
				\hline	
				
				\multirow{3}{*}{3l-pointnet++} & $C$ & 512 &  128 & 1 & - &  & \multirow{3}{*}{max} \\
				&  $k$ &  64   & 64  & 128  & - & pointMLP \\
				&  $m$ &  128   & 256  & 1024  & - &  \\
				\hline			
				\multirow{3}{*}{4l-pointnet++} & $C$ &  512 & 128 & 32 & 1 & & \multirow{3}{*}{max} \\
				&  $k$ & 32   & 32  & 8  & 32 & pointMLP   \\
				&  $m$ & 128   & 256  & 512  & 1024 &    \\
				\hline	
				\hline	
				\multirow{3}{*}{4l-spec-max} & $C$ &   512 & 128 & 32 & 1 & & \multirow{3}{*}{max} \\
				&  $k$ & 32   & 32  & 8  & 32 & spec-conv  \\
				&  $m$ & 128   & 256  & 512  & 1024 &    \\
				\hline	
				\multirow{3}{*}{4l-spec-cp} & $C$ &   512 & 128 & 32 & 1 & &\multirow{3}{*}{cp} \\
				&  $k$ & 32   & 32  & 8  & 32 & spec-conv \\
				&  $m$ & 128   & 256  & 512  & 1024 &    \\
				\hline	
				
			\end{tabular}
		}
	}
	\hfill
	\parbox{.5\linewidth}{
		\centering
		\small
		\renewcommand{\arraystretch}{1.2}
		\scalebox{0.80}{
			\begin{tabular}{l c | c c c c |c|c}
				\hline
				
				\multicolumn{2}{l|}{Structure}	& L1  			& L2		   & L3 		& L4	&	Kernel    & Pooling 		  \\
				\hline	
				
				\multirow{3}{*}{3l-pointnet++} & $C$ & 1024 &  256 & 1 & - &  & \multirow{3}{*}{max} \\
				&  $k$ &  64   & 64  & 256  & -& pointMLP \\
				&  $m$ & 128   & 256  & 1024  & - &    \\
				\hline			
				\multirow{3}{*}{4l-pointnet++} & $C$ &  1024 & 256 & 64 & 1 & & \multirow{3}{*}{max} \\
				&  $k$ & 32   & 32  & 8  & 64 & pointMLP   \\
				&  $m$ & 128   & 256  & 512  & 1024 &    \\
				\hline	
				\hline	
				\multirow{3}{*}{4l-spec-max} & $C$ &  1024 & 256 & 64 & 1 & & \multirow{3}{*}{max} \\
				&  $k$ & 32   & 32  & 8  & 64 & spec-conv  \\
				&  $m$ & 128   & 256  & 512  & 1024 &    \\
				\hline	
				\multirow{3}{*}{4l-spec-cp} & $C$ &  1024 & 256 & 64 & 1 & &\multirow{3}{*}{cp} \\
				&  $k$ & 32   & 32  & 8  & 64 & spec-conv \\
				&  $m$ & 128   & 256  & 512  & 1024 &    \\
				\hline	
				
			\end{tabular}
		}
	}
	\vspace{0.4cm}
	\caption{Network architectures for the 1k experiments (left) and the 2k experiments (right). Here, for each layer, $C$ stands for the number of centroids, $k$ stands for the size of the $k$-NN, and $m$ stands for the output feature dimension.}
	\vspace{-1.0cm} 
	\label{tab:Architecture}
\end{table}

\subsection{Network Architecture and Training}\label{sec:net_arch}
We provide details of our network structures for the case of 1024 input points (1k) and 2048 input points (2k). The network structure for the 2k experiments is designed to be ``wider'' to better accommodate the increased input point density. Table \ref{tab:Architecture} lists all the variations of the pointnet++ and our spectral point convolution network structures, which we will later to refer to when presenting experimental results. The 3l-pointnet++ is that defined in the ``pointnet2\_cls\_ssg.py'' model file on the pointnet++ GitHub page \footnote{\small \url{https://github.com/charlesq34/pointnet2/blob/master/models/pointnet2_cls_ssg.py}}. We replace the kernels from 4l-pointnet++ with spectral graph convolution kernels to acquire the 4l-spec-max model. Replacing max pooling with recursive cluster pooling in the 4l-spec-max model results in the 4l-spec-cp model. 

Configurations for the layers after L3/L4 in Table \ref{tab:Architecture} differ for the classification and segmentation tasks. For classification on the McGill Shape Benchmark, ModelNet40 and MNIST, we used 3 fully connected layers with drop out, i.e., FC(512, 0.5) $\rightarrow$ FC(256, 0.5) $\rightarrow$  FC(\#classes). For segmentation on the ShapeNet Part Segmentation and ScanNet datasets, feature propagation (FP) layers (as in \cite{qi2017pointnet}) are applied after L3/L4. The number of FP layers and their input/output dimensions are the same as those of the corresponding previous set activation layers. After the FP layers, two fully connected layers are applied to map learned features to point labels. 

In all our experiments, we applied the following strategy for network training. We used the Adam optimizer \cite{kingma2014adam}
with an initial learning rate of 0.001 and an exponential decay on the learning rate with a ratio of 0.5 applied every 20 epochs. We chose to use the Relu activation function and applied batch normalization (BN) with a size of 32 and a decay rate set to increase from 0.5 to 0.99. Throughout we followed the strategy in pointnet++ \cite{qi2017pointnet} for data augmentation: random rotation around the up-right direction, small rotation perturbation around the $x,y,z$ axes, point location perturbation, and random scaling and translation.

\subsection{Ablation Study for Network Models}\label{sec:modelAblation}
We now evaluate the effect of the novel components in our framework: 1) local spectral filtering on a $k$-NN and 2) recursive cluster pooling for local outputs. We apply a 4 layer pointnet++ structure as the baseline method, then add spectral filters to each layer, and then replace max pooling with recursive cluster pooling. We also include results obtained using the 3 layer structure used in \cite{qi2017pointnet}.  In addition, we consider the scalability of both approaches, by varying the number of input points, the effect of including additional features such as surface normals, and training time. These results are presented in Table (\ref{tab:Details}). 

\begin{table}[!t]
\centering
\scalebox{1.0}{
\small
\renewcommand{\arraystretch}{1.2}
\hspace{-0.3cm}
\begin{tabular}{l|c|c|c|c|c}
\hline
ModelNet40   & Acc 1k  & Acc 1k +N & Time 250ep & Acc 2k  +N & Time 250ep \\
\hline
3l-pointnet++              & 90.7     &  91.3   & 11h  &   91.5  &  20h             \\
4l-pointnet++         & 90.6     &   91.1    & 7.5h             &   91.2    &      11h        \\
\hline
4l-spec-max          & 91.2      &  91.6   & 8h                & 91.9        &  12h         \\
4l-spec-cp &  91.5     &   91.8   & 12h                & \cellcolor{pink} 92.1       &   20h        \\

\hline 
\hline
ShapeNet Seg           &  mIOU 1k     &  mIOU 1k +N    &    Time  100ep         &    mIOU 2k +N     &  Time 100ep       \\
\hline
3l-pointnet++         & 84.2      &  84.7 &        7.5h          & 84.9       &      14h    \\
4l-spec-cp        & 84.6       &  85.0  &           8h     & \cellcolor{pink}85.4        &      14h      \\
\hline 
\end{tabular}
}
\vspace{0.3cm}
\caption{{\bf Model Ablation Study on ModelNet40 (classification) and ShapeNet (segmentation).} Acc stands for classification accuracy, 1k/2k refers to the number of points used and ``+N'' indicates the addition of surface normal features to xyz. For the segmentation experiments, mIOU stands for mean intersection over union. Here we only compare the best models from pointnet++ with ours. Training time is with respect to the number of epochs used in each experiment. Adding normals only increases the training time by a negligible amount, therefore only one runtime column is provided for the 1k experiments. Network structures for all the reported experiments are in Table \ref{tab:Architecture}.}
\label{tab:Details}
\vspace{-0.5cm}
\end{table}

From the model ablation study in Table \ref{tab:Details}, it is evident that our proposed model, which incorporates spectral graph convolution together with recursive cluster pooling, provides a non-trivial improvement over pointnet++ on both the classification and segmentation tasks. We make the following observations:
1) {\em 3l-pointnet++ performs better than the 4 layer version.} This is likely because in the 4 layer version the neighborhood size is half of that in the 3 layer version. Since features are learned at each point in an isolated fashion in pointnet++, the use of larger neighborhoods gives an advantage.
2) {\em Spectral graph convolution on local $k$-NNs performs better than point-wise MLP.} The 4l-spec-max model outperforms 4l-pointnet++. This implies that the topological information encoded by spectral graph convolution benefits feature learning.
3) {\em Recursive cluster pooling further boosts the performance of the spectral graph convolution layer.} This suggests that information aggregation following spectral coordinates increases the discriminative power of the learned point features, benefiting both classification and segmentation.
4) {\em The runtime of our model is comparable to those of pointnet++.} The eigenvalue decomposition used in spectral convolution and recursive cluster pooling could in theory be costly, but since we use local neighborhoods the impact is not severe. Our best model, 4l-spec-cp, has roughly the same training time as that of 3l-pointnet++, which is the best model from pointnet++. Spectral graph convolution kernels are as fast as the point-wise MLP kernels, which can be seen by comparing the runtime of the 4l-spec-max and 4l-pointnet++ models.

We now provide comparisons against the present state-of-the-art methods, in both classification and segmentation tasks, on
various datasets described in Section \ref{sec:datasets}. When comparing against pointnet++, unless stated otherwise, we apply the 3l-pointnet++ model since it gives better results than the 4 layer version in our model ablation study in Table \ref{tab:Details}.

\subsection{Classification Experiments}

\subsubsection{McGill Shape Benchmark}\label{sec:exp_mcgill}

The classification results for the McGill Shape Benchmark are presented in Table \ref{tab:mcgillAblation}, using 1024 $xyz$ points as the inputs in all cases. Spectral graph convolution on point sets provides a consistent boost in both average instance level accuracy and category level accuracy. 
Further, the use of recursive cluster pooling grants our model another $0.7\%$ boost in overall instance level accuracy over max pooling. 
Since the $k$-NNs may contain disjoint sets of points, a recursive aggregation of $k$-NN features by clustering the spectral coordinates appears to increase discriminative power for articulated objects.

\begin{table}[h!]
\footnotesize
\centering
\scalebox{0.9}{
  \begin{tabular}{llccccccc}  
    \toprule
    & & &\multicolumn{3}{c}{Avg Instance Level Accuracy (\%) } & \multicolumn{3}{c}{Avg Category Level Accuracy (\%)} \\
    \cmidrule(lr){4-6} \cmidrule(lr){7-9}
  \multirow{2}{*}{ Model}  & \multirow{2}{*}{Kernel}& \multirow{2}{*}{Pooling} & \multirow{2}{*}{Articulated}  &  \multirow{2}{*}{Non-Arti} & \multirow{2}{*}{Combined} &  \multirow{2}{*}{Articulated}  &  \multirow{2}{*}{Non-Arti} & \multirow{2}{*}{Combined} \\
     &  &  &  &  &  &  & &  \\
    \cmidrule(lr){1-3} \cmidrule(lr){4-6} \cmidrule(lr){7-9}
    3l-pointnet++ & point-MLP & max  & 91.25 & 95.31 & 93.06 & 91.33 & 95.44 & 93.27 \\
    \cmidrule(lr){1-3} \cmidrule(lr){4-6} \cmidrule(lr){7-9}
    4l-pointnet++ & point-MLP & max  & 92.50 & 92.19 & 92.36 & 92.83 & 92.74 & 92.79  \\
    \cmidrule(lr){1-3} \cmidrule(lr){4-6} \cmidrule(lr){7-9}
   	\multirow{2}{*}{4l-spec-cp} & spec-conv & max  & 92.50 & 98.44 & 95.14 & 92.75 & 98.41 & 95.43  \\
    & spec-conv & cp & 93.75 & 98.44 & \cellcolor{pink} 95.83 & 93.30 & 98.41 & \cellcolor{pink} 95.74  \\
    \bottomrule
  \end{tabular}
  }
  \vspace{0.2cm}
  \caption{McGill Shape Benchmark classification results. We report the instance and category level accuracy on both the entire database and on subsets (see Table \ref{tab:Architecture} for network structures).}
  \label{tab:mcgillAblation}
  \vspace{-1.0cm}
\end{table}

\subsubsection{MNIST dataset}\label{sec:exp_mnist}
2D images can be treated as a grid graph \cite{defferrard2016convolutional,monti2016geometric} or a 2D point cloud \cite{qi2017pointnet}. We provide results on the MNIST dataset using our proposed best model, 4l-spec-cp, from the previous model ablation study. We compare our results with the state-of-the-art methods in graph CNNs \cite{defferrard2016convolutional,monti2016geometric}, in point sets \cite{qi2017pointnet} \footnote{We tried to reproduce the pointnet++ results on MNIST. Our $0.55 \%$ error rate is very close to the $0.5 \%$ error rate reported by the authors.} and with regular neural network/CNN approaches applied on the 2D image domain \cite{simard2003best,lecun1998gradient,lin2013network}.

For both pointnet++ and our method, 784 points are provided as inputs to the network and we use the 1k experimental network, where the first layer samples 512 centroids (see Table \ref{tab:Architecture}).
The results in Table \ref{tab:MNIST} show that approaches which favor local operations on the input domain usually yield better performance, for instance, MLP vs. LeNet, and our method vs. ChebNet. Our approach gives a 20\% error rate reduction over pointnet++, demonstrating the advantage of spectral convolution on a local $k$-NN graph over the isolated learning process in point-wise MLP. In addition, our performance surpasses that of the Network in Network model \cite{lin2013network}, which is a strong regular image CNN model.

\begin{table*}[h]
	\renewcommand{\arraystretch}{1.2}
	\centering
	\small
	\vspace{-0.5cm}
	\begin{tabular}{@{}l|c|c|c}
		\hline
		Method & Domain & Kernel & Error Rate(\%) \\ \hline
		Multi-layer perceptron \cite{simard2003best} & full image & spatial MLP& 1.60  \\
		LeNet5 \cite{lecun1998gradient} & local img patch & spatial conv & 0.80  \\
		Network in Network \cite{lin2013network} & local img patch & spatial conv & 0.47 \\ \hline
		ChebNet \cite{defferrard2016convolutional} & full graph & spectral graph conv & 0.86 \\
		MoNet \cite{monti2016geometric} & local graph & spatial graph conv & 0.81 \\ \hline
		3l-pointnet++ \cite{qi2017pointnet} & local points & spatial point-MLP &  0.55 \\
		4l-spec-cp & local $k$-NN graph & spectral graph conv & \cellcolor{pink} 0.42  \\ \hline
	\end{tabular}
	\vspace{0.2cm}
	\caption{Results on the MNIST dataset. For the pointnet++ results, we reproduced their experiments, as discussed in \cite{qi2017pointnet}.}
	\label{tab:MNIST}
	\vspace{-1.6cm}
\end{table*}

\begin{table}[t]
\renewcommand{\arraystretch}{1.2}
    \centering
        \small
 	\begin{tabular}[width=\linewidth]{l|c|c|c|c|cc}
        \hline
        Method  & Domain & Kernel & Pooling & Acc (\%) & Acc + N (\%) \\
        \hline
        Subvolume~\cite{qi2016volumetric} &Voxel Grid & 3D conv & 3D-max  & 89.2  &  - \\
        MVCNN~\cite{su2015multi} & 2D views & 2D conv & view-max  & 90.1 &  - \\
        PointNet~\cite{qi2016pointnet} & Full Points& point-MLP &point-max & 89.2 &  -\\ 
        \hline
        \hline
        3l-pointnet++ \cite{qi2017pointnet} & Local Points & point-MLP & point-max  & 90.7 &  91.5\\ 
        4l-spec-max & Local $k$-NN graph & graph conv & graph-max  &  91.2 &  91.9\\
        4l-spec-cp & Local $k$-NN graph & graph conv & graph-cp & 91.5 &  \cellcolor{pink} 92.1\\
        \hline
    \end{tabular}
    
    \vspace{0.2cm}
    \caption{ModelNet40 results. ``Acc'' stands for 1k experiments with only $xyz$ points as input features. ``Acc + N'' stands for 2k experiments with $xyz$ points along with their surface normals as input features.``graph-cp'' stands for recursive cluster pooling, introduced in section \ref{sec:specpooling}.
    }
    \label{tab:modelnet40}
    \vspace{-.5cm}
 \end{table}

\subsubsection{ModelNet 40 Dataset}\label{sec:exp_modelnet}
We present ModelNet40 3D shape recognition results in Table \ref{tab:modelnet40}, where we compare our method with representative competitive approaches. We were able to reproduce the results from pointnet++, to get very similar performance to that reported by the authors in \cite{qi2017pointnet}. We report two sets of accuracy results. In the first 1024 $xyz$ point coordinates are used as inputs, with the network structure following the 1k configurations in Table \ref{tab:Architecture}. In the second 2048 $xyz$ points along with their surface normals are used as inputs, with the network structure following the 2k configurations in Table \ref{tab:Architecture}. Our use of spectral graph convolution and recursive cluster pooling provides a consistent improvement over pointnet++, and leads to state-of-the-art level classification performance on the ModelNet40 Benchmark.

\subsection{Segmentation Experiments}
Point segmentation is a labeling task for each point in a 3D point set, and is more challenging than point set classification. We present experimental results on ShapeNet \cite{yi2016scalable} and ScanNet \cite{dai2017scannet} in Table \ref{tab:seg}.
We then provide details on experimental settings for each case.

\begin{table}[h]
    \centering
        \small
        \renewcommand{\arraystretch}{1.2}
 	\begin{tabular}[width=\linewidth]{l|c|c|c|c}
        \hline
        Method & Domain & Kernel  & ShapeNet-mIOU (\%) & ScanNet-Acc (\%)  \\
        \hline
        Yi \etal \cite{yi2016scalable} & - & - & 81.4 & - \\
        Dai \etal \cite{dai2017scannet} & Voxel Grid & 3D conv & - & 73.0 \\
        SynSpecCNN \cite{yi2017syncspeccnn} & Full $k$-NN Graph & graph conv &  84.7 & - \\
        PointNet~\cite{qi2016pointnet} & Full Points & point-MLP & 83.7 & 73.9  \\ 
        \hline
        \hline
        Pointnet++ \cite{qi2017pointnet} & Local Points & point-MLP &  84.9 & 84.0 \\
        4l-spec-cp & Local $k$-NN Graphs & graph conv &  \cellcolor{pink}  85.4 & \cellcolor{pink} 84.8 \\
        \hline
    \end{tabular}
    \vspace{0.2cm}
    \caption{A comparison between our method and the present state-of-the-art approaches in segmentation tasks. For ShapeNet, mIOU stands for mean intersection over union on points, and for ScanNet, Acc stands for voxel label prediction accuracy.} 
    \label{tab:seg}
     \vspace{-.5cm}
 \end{table}

\subsubsection{Shapenet Part Segmentation Dataset}\label{sec:exp_shapenet}
We compare our method with state-of-the-art approaches, as well as the reproduced results from pointnet++. Following the setting in \cite{yi2016scalable}, we evaluate our approach assuming that a category label for each shape is already known and we use the same mIoU (mean intersection over union) metric on points. 2048 $xyz$ points and their surface normals are used as input features and the network structure follows that of the 2k configurations in Table \ref{tab:Architecture}. More specifically, 3l-pointnet++ model is applied for pointnet++ and 4l-spec-cp is applied for our method.

\subsubsection{ScanNet Dataset}
ScanNet is a large-scale semantic segmentation dataset constructed from real-world 3D scans of indoor scenes, and as such is more challenging than the synthesized 3D models in ShapeNet. Following \cite{qi2017pointnet}\cite{dai2017scannet}, we remove RGB information in our experiments in Table \ref{tab:seg} and we use the semantic voxel label prediction accuracy for evaluation. The training and testing procedures follow those in pointnet++ \cite{qi2017pointnet}. 8192 $xyz$ points are used as input features and the network structure is that of the 2k configurations in Table \ref{tab:Architecture}. More specifically, the 4l-pointnet++ model is applied for pointnet++ and the 4l-spec-cp is applied for our method. \footnote{The 3l-pointnet++ model leads to inferior performance on this large-scale indoor dataset. For both networks, for all experiments reported in this paper, single scale grouping (SSG in \cite{qi2017pointnet}) is applied for a fair comparison.
}

\subsubsection{Results and Discussion}
\begin{table}[t]
	\begin{center}
		\begin{tabular}{m{0.7cm} m{1.8cm} m{2.cm} m{2cm} m{1.8cm} m{1.8cm} m{1.8cm}}
			& \hspace{0.1cm}\textbf{ handbag} & \textbf{skate-board} & \hspace{0.3cm} \textbf{table} & \textbf{lamp} & \hspace{-0.25cm}\textbf{airplane}& \textbf{knife}\\	
			\textbf{PN++}  &			\trimmedgraphic{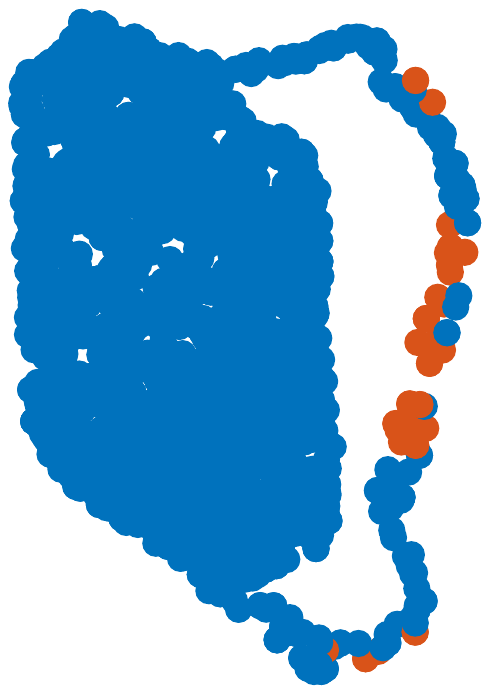} &			\trimmedgraphic{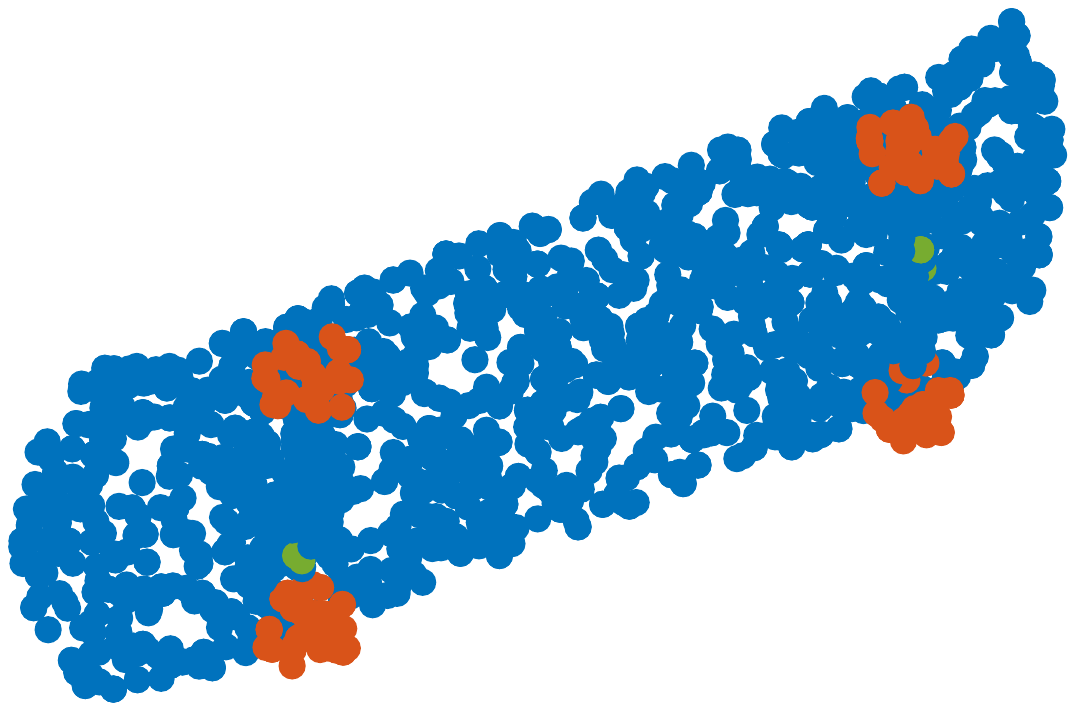} & 	\trimmedgraphic{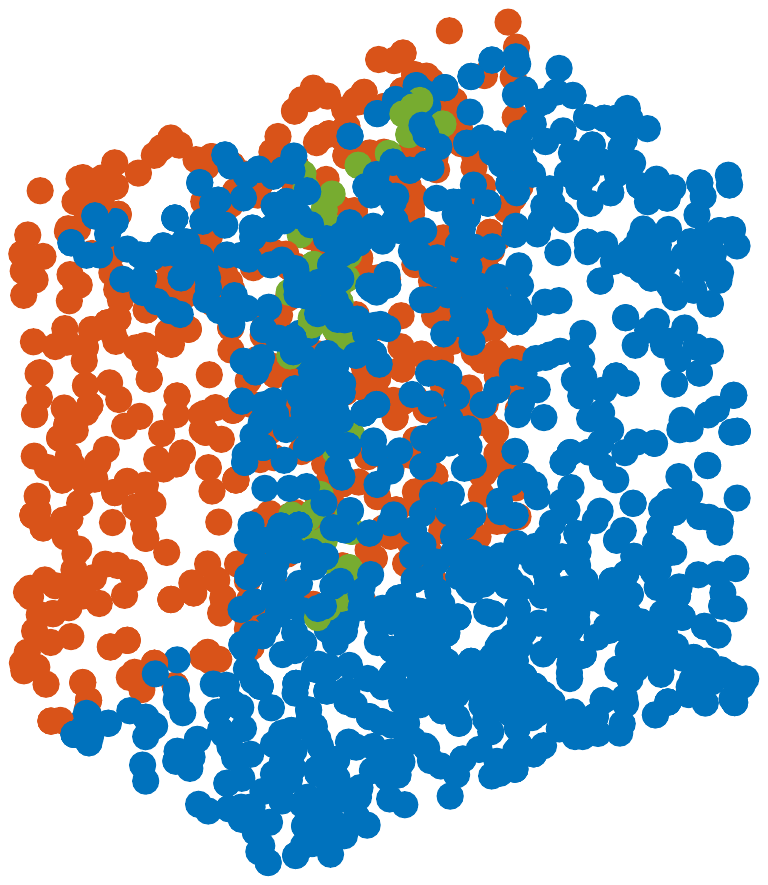}&\hspace{-0.25cm} \trimmedgraphic{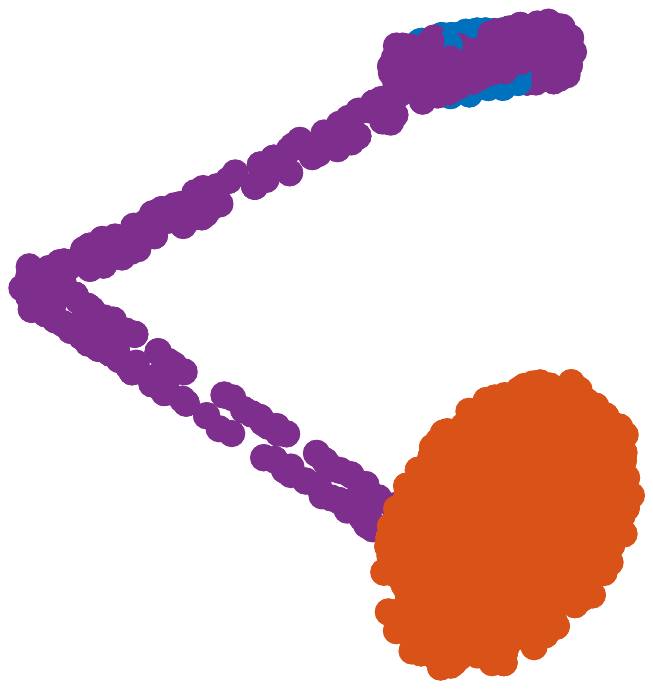}&\hspace{-0.5cm}\trimmedgraphic{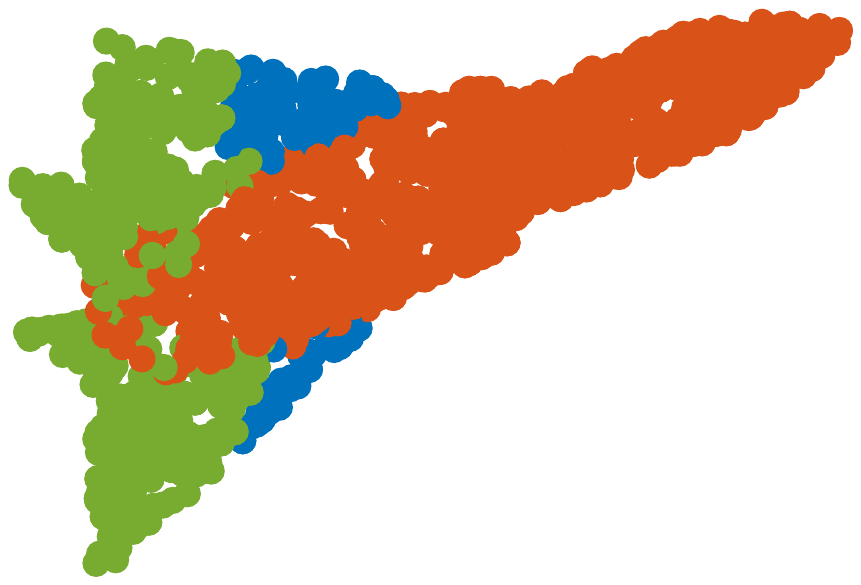}&\hspace{-0.4cm}\trimmedgraphic{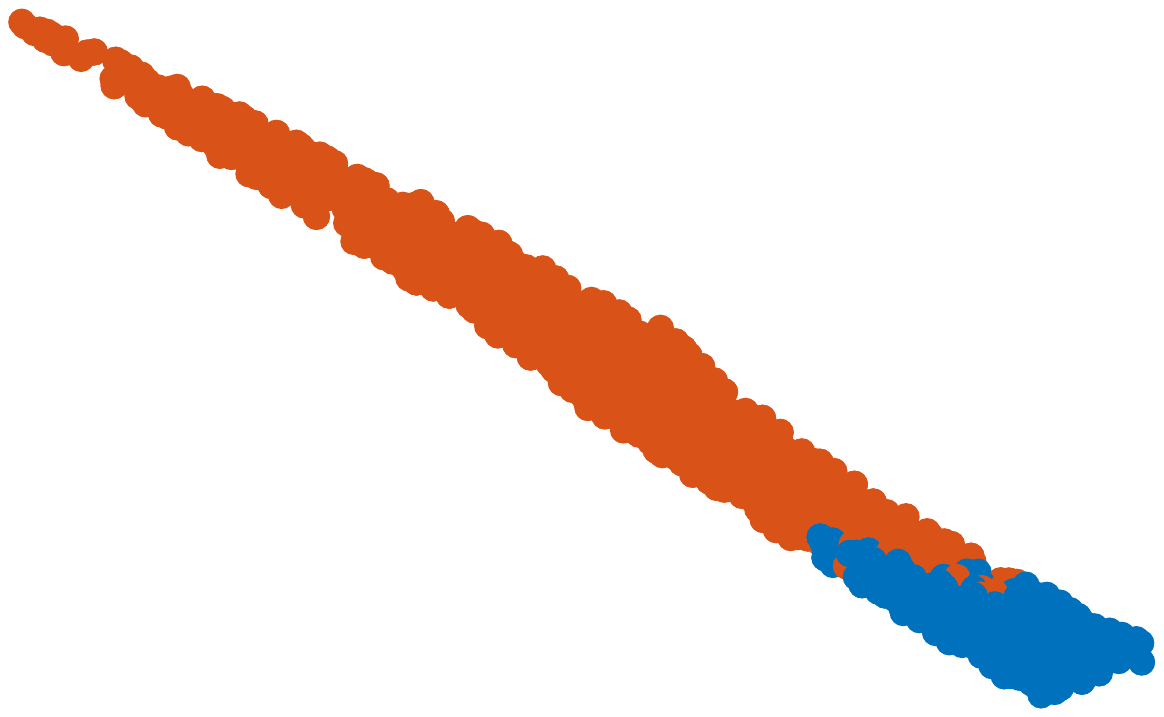}\\
			
			\textbf{Ours}  &			\trimmedgraphic{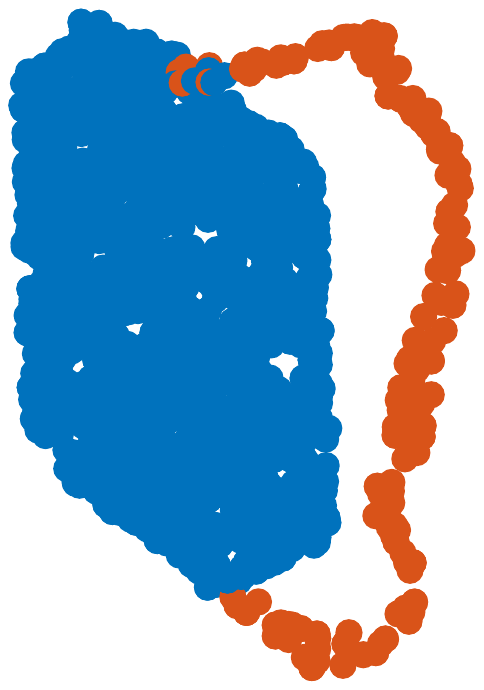} &			\trimmedgraphic{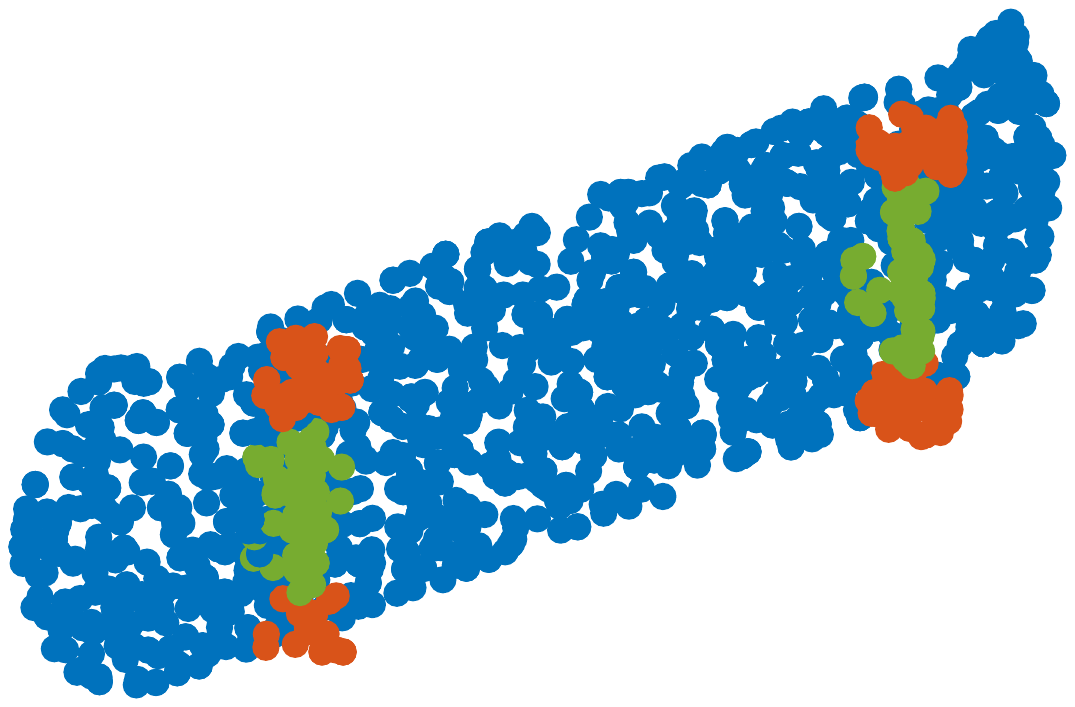} & 	\trimmedgraphic{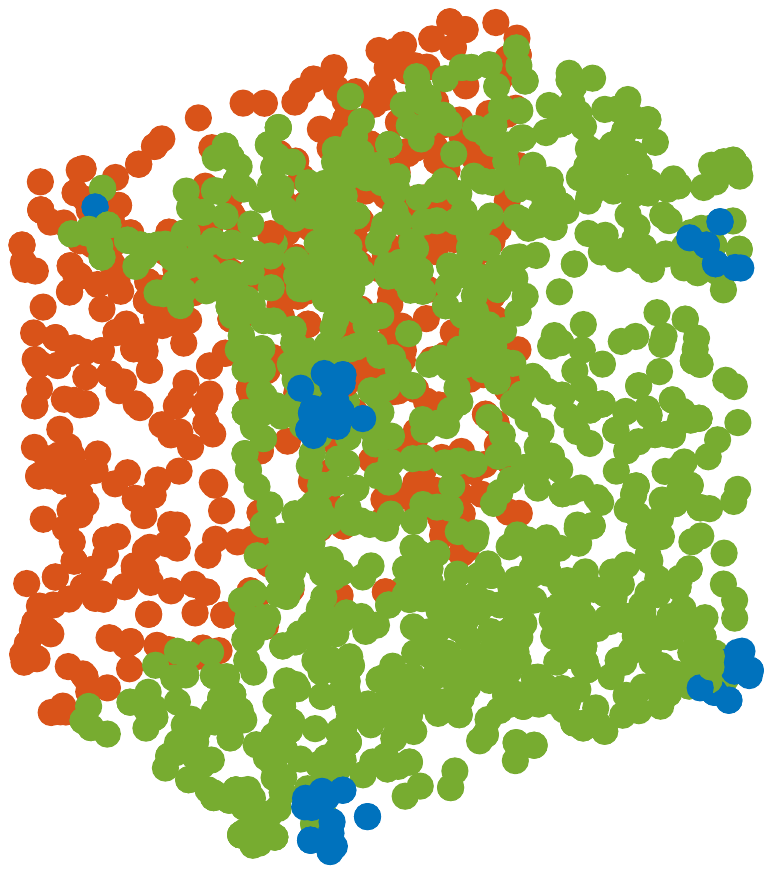}&\hspace{-0.25cm} \trimmedgraphic{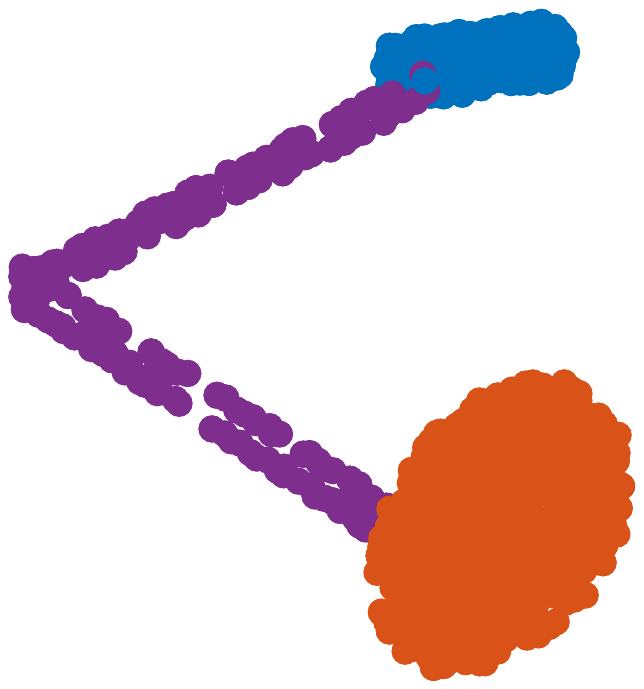} &\hspace{-0.5cm}	\trimmedgraphic{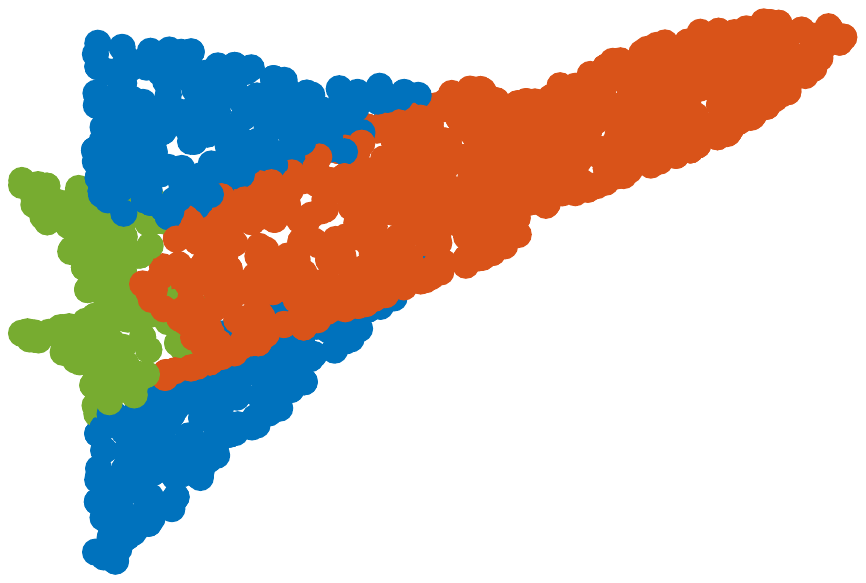}&\hspace{-0.4cm}	\trimmedgraphic{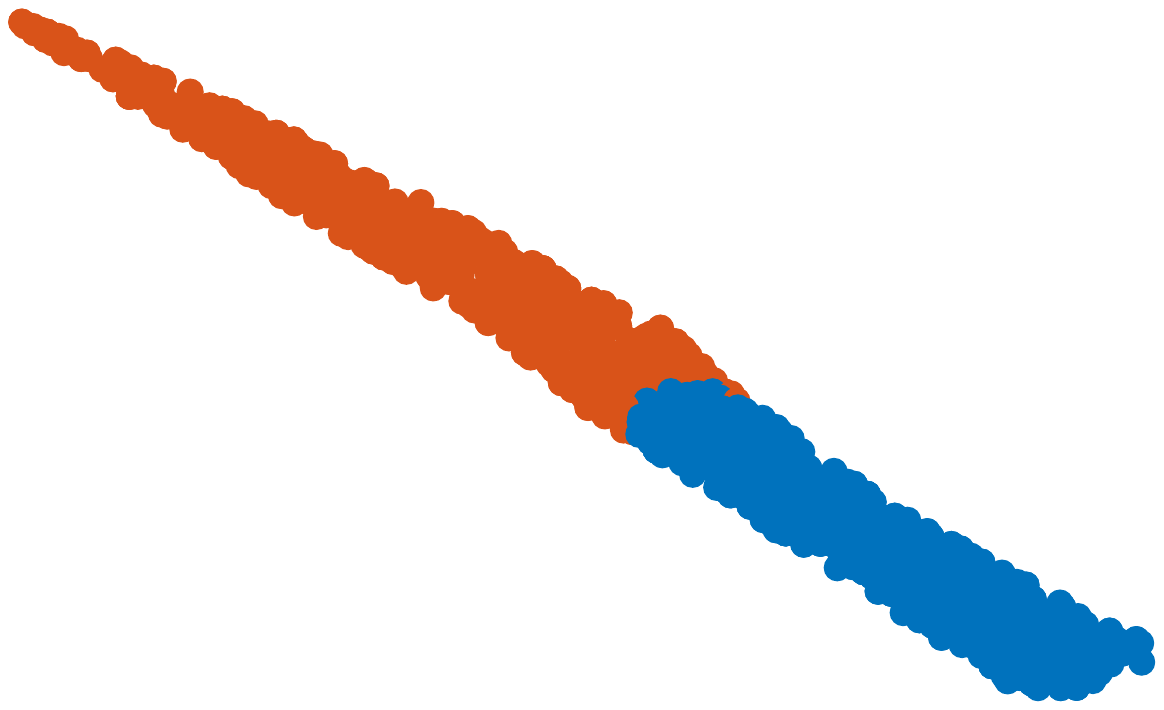}\\
			
			\textbf{GT} &			\trimmedgraphic{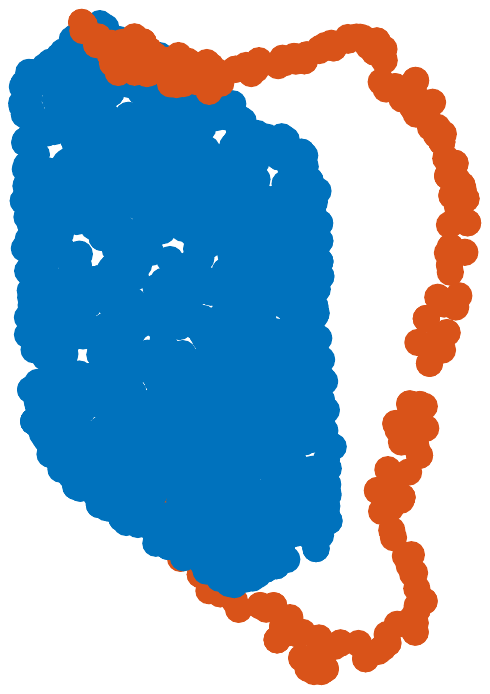} &			\trimmedgraphic{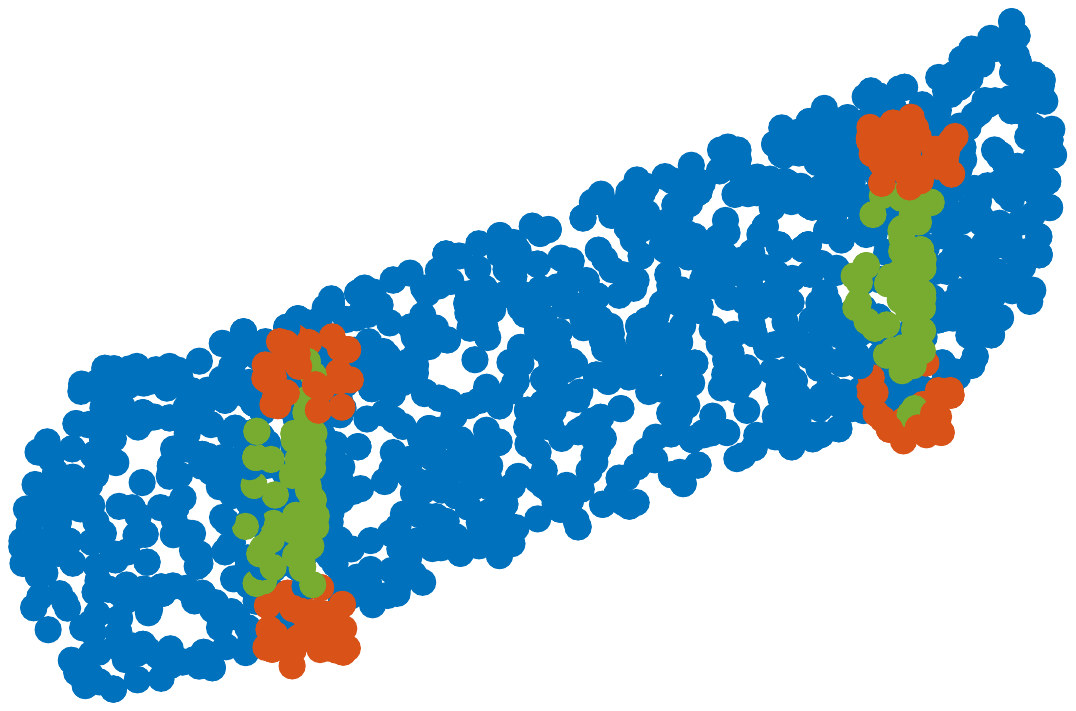} & 	\trimmedgraphic{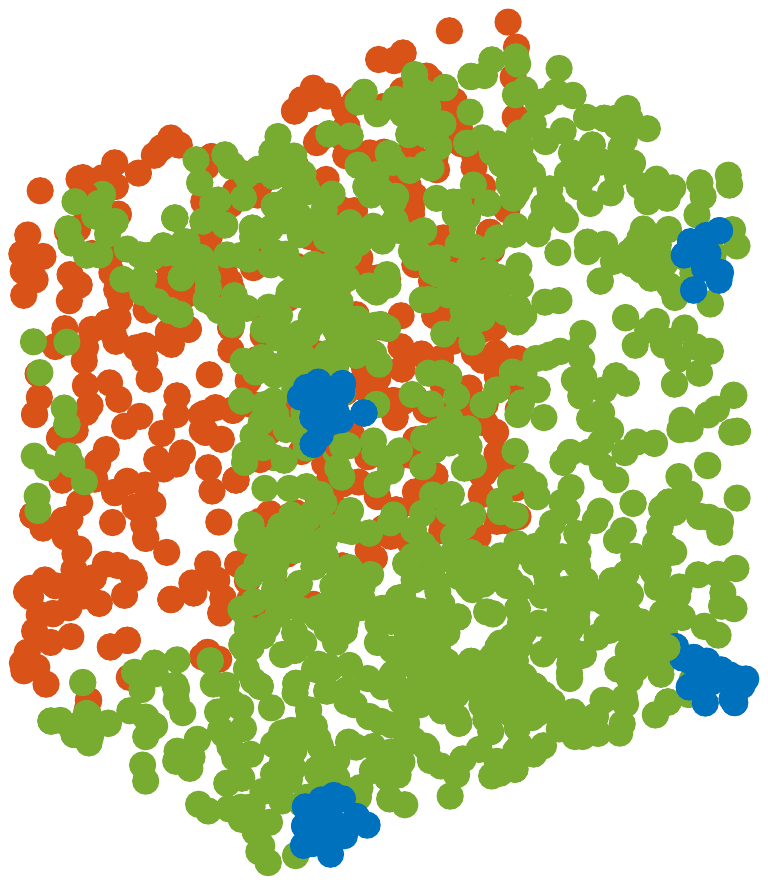}&\hspace{-0.25cm} \trimmedgraphic{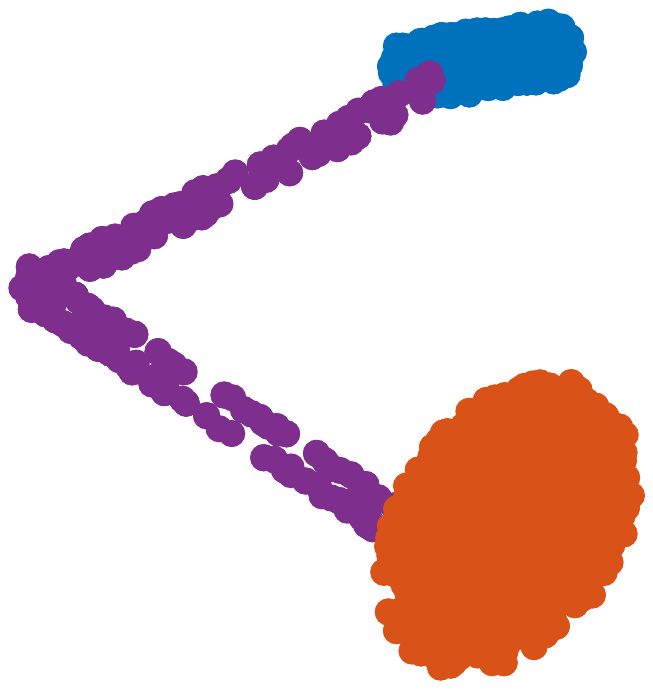}&\hspace{-0.5cm}	\trimmedgraphic{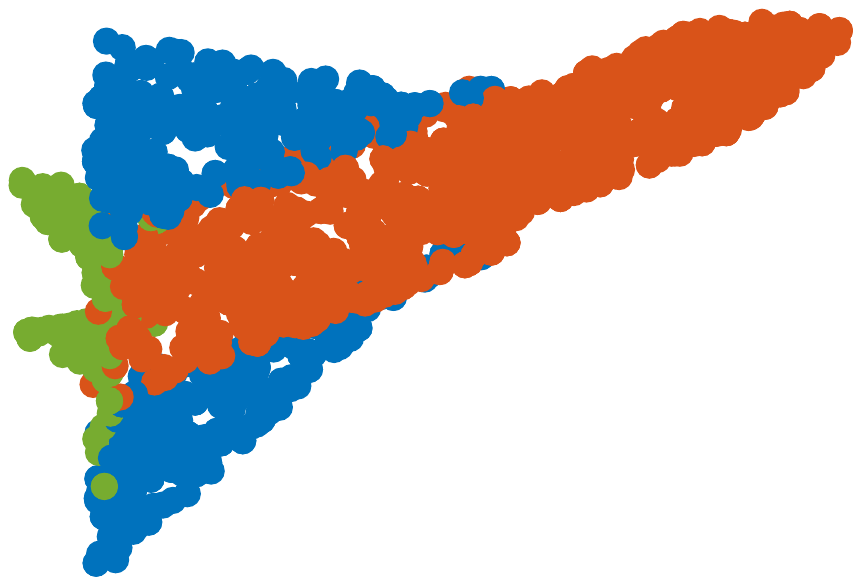}&\hspace{-0.4cm}	\trimmedgraphic{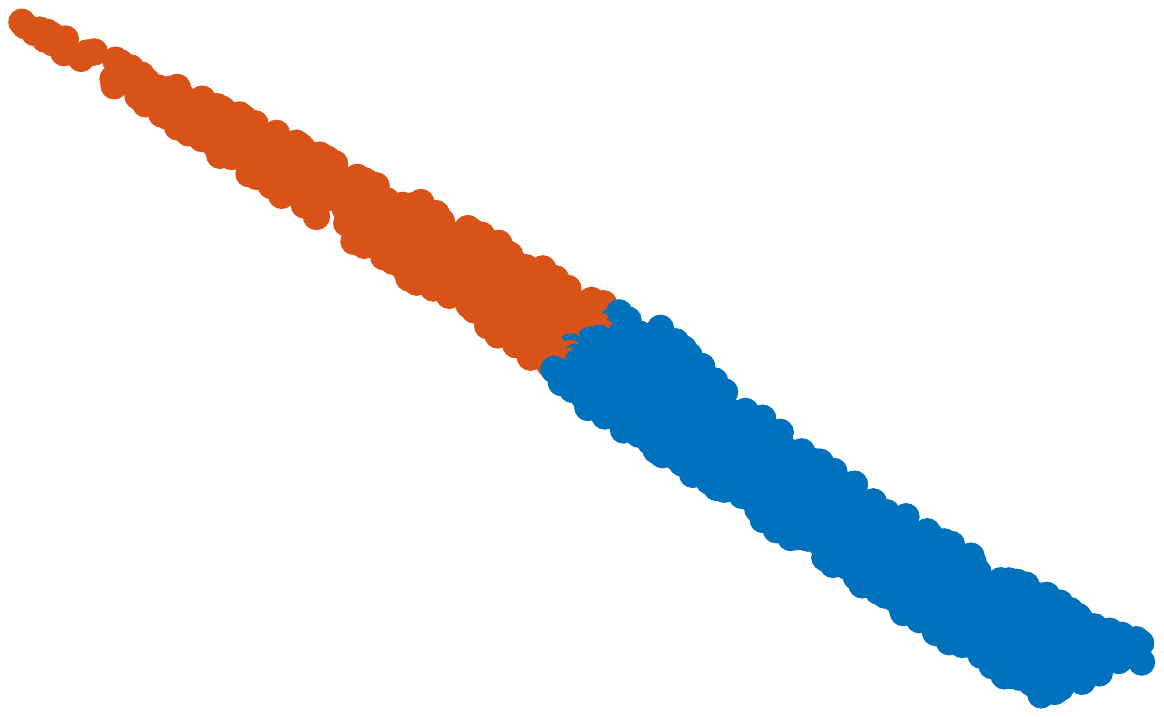}\\
			
		\end{tabular}
	\end{center}
	\vspace{-0.4cm}
	\captionof{figure}{A visualization of segmentation results on ShapeNet using pointnet++ (top row, 84.9\% in Table \ref{tab:seg}), our 41-spec-cp model (middle row, 85.4\% in Table \ref{tab:seg}), and the ground truth labels (bottom row). Our method appears to better capture fine local structures (see text for a discussion).}
	\label{fig:seg_res}
	\vspace{-0.8cm}
\end{table}

The use of spectral graph convolution combined with cluster pooling in our approach once again provides a non-trivial improvement over pointnet++, achieving state-of-the-art level performance on both part segmentation (ShapeNet) and indoor scene semantic segmentation (ScanNet). 
We provide an illustrative visualization of the part segmentation results on selected models in Fig. \ref{fig:seg_res}. In these examples, when compared to pointnet++, our approach gives results that are closer to the ground truth overall and better captures fine local structures, such as the axles of the skateboard, and the feet of the table. In addition, spectral graph convolution with cluster pooling provides a more faithful representation of changes in local geometry. This allows us to successfully segment connected parts of a 3D object, such as the strap from the body of the hand bag, the wings from the tail fins of the airplane and the handle from the blade of the knife.

\section{Conclusion}
The use of spectral graph convolution on local point neighborhoods, followed by recursive cluster pooling on the resultant representations, holds great promise for feature learning from unorganized 3D point sets. Our method's ability to capture local structural information and geometric cues presents an advance in deep learning approaches to feature abstraction from unorganized point sets in 3D computer vision. Considering its strong experimental performance, acceptable runtime, and versatility in handling a variety of datasets and tasks, our approach could have considerable practical value as 3D depth sensors begin to become more and more common place.

\clearpage

\bibliographystyle{splncs03}
\bibliography{ref}
\end{document}